\documentclass{bmvc2k}
\usepackage{bm, multirow, amsmath, amssymb, array, floatrow}

\title{Probabilistic Estimation of 3D Human Shape and Pose with a Semantic Local Parametric Model}

\addauthor{Akash Sengupta}{as2562@cam.ac.uk}{1}
\addauthor{Ignas Budvytis}{ib255@cam.ac.uk}{1}
\addauthor{Roberto Cipolla}{rc10001@cam.ac.uk}{1}

\addinstitution{
 Department of Engineering\\
 University of Cambridge\\
 Cambridge, UK
}

\runninghead{Sengupta \etal}{Probabilistic Human Shape \& Pose with a Local Model}

\def\eg{\emph{e.g}\bmvaOneDot}

\def\etal{\emph{et al}\bmvaOneDot}

\begin{document}

\maketitle

\begin{abstract}
This paper addresses the problem of 3D human body shape and pose estimation from RGB images. Some recent approaches to this task predict probability distributions over human body model parameters conditioned on the input images. This is motivated by the ill-posed nature of the problem wherein multiple 3D reconstructions may match the image evidence, particularly when some parts of the body are locally occluded. However, body shape parameters in widely-used body models (\eg SMPL) control global deformations over the whole body surface. Distributions over these \textit{global} shape parameters are unable to meaningfully capture uncertainty in shape estimates associated with \textit{locally}-occluded body parts. In contrast, we present a method that (i) predicts distributions over local body shape in the form of semantic body \textit{measurements} and (ii) uses a linear mapping to transform a local distribution over body measurements to a global distribution over SMPL shape parameters. We show that our method outperforms the current state-of-the-art in terms of identity-dependent body shape estimation accuracy on the SSP-3D dataset, and a private dataset of tape-measured humans, by probabilistically-combining local body measurement distributions predicted from multiple images of a subject.
\end{abstract}

\section{Introduction}

\begin{figure}
    \centering
    \includegraphics[width=0.975\textwidth]{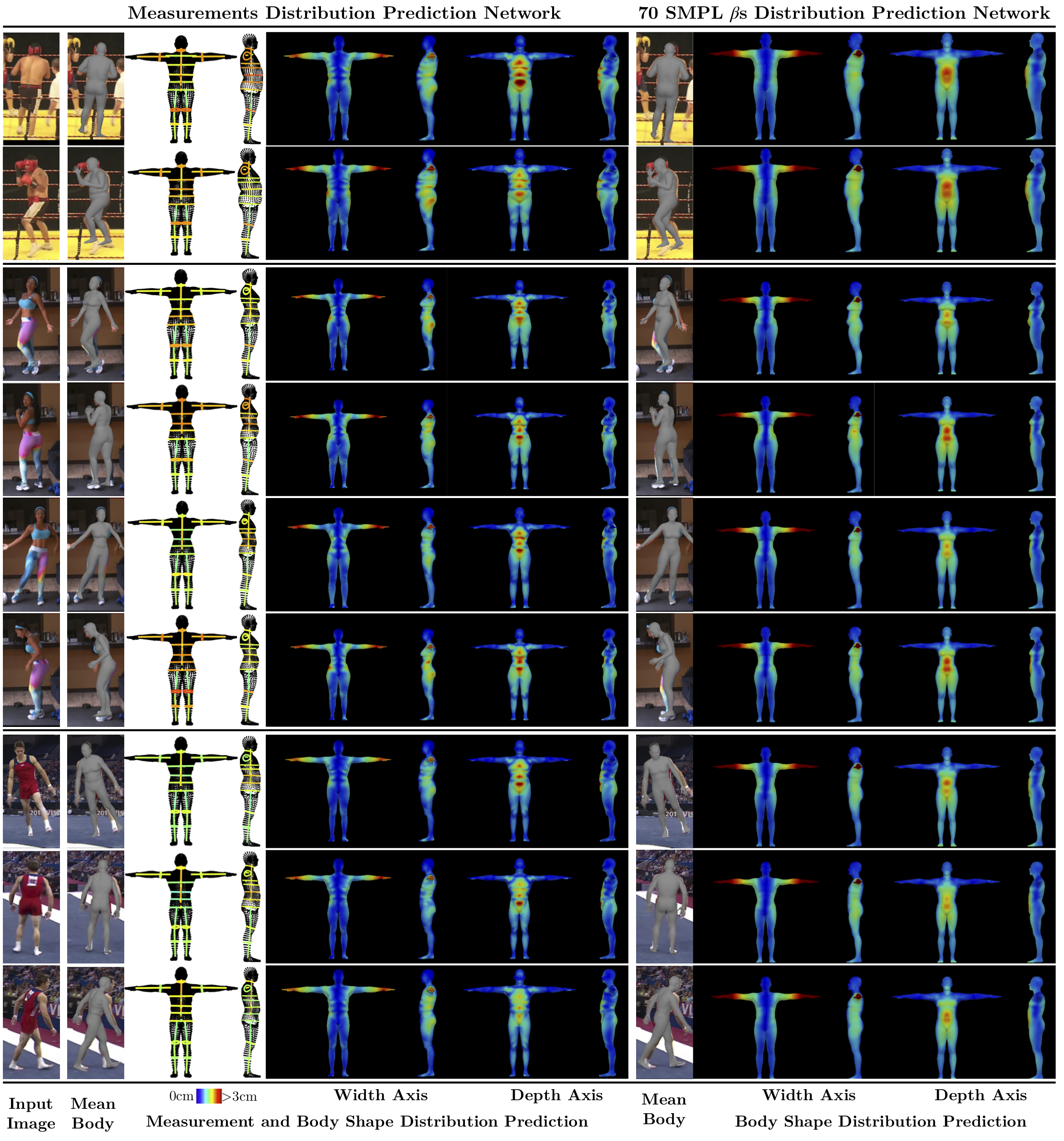}
    \caption{Comparing independent Gaussian measurement distributions and SMPL $\beta$ distributions on images from SSP-3D \cite{STRAPS2020BMVC}. Measurement distributions predictions (left) exhibit meaningful \textit{local} shape uncertainty arising from varying camera angles, challenging poses and self-occlusions. For example, comparing rows 1 vs 2 shows that front/back-facing images result in larger predicted uncertainty (i.e. variance) for \textit{depth} measurements (columns 4, 7, 8), while side-facing images result in greater uncertainty for \textit{width} measurements (columns 3, 5, 6). This is reasonable as body depth is ambiguous from a front-on viewpoint while body width is ambiguous from the side. Moreover, in row 3 the subject's hips are obscured by their pose but the upper torso is visible, while the opposite is true in row 4 where hair occludes the torso. Accordingly, row 3 shows larger hip width uncertainty while row 4 shows larger torso width uncertainty (columns 3 and 6). In contrast, independent Gaussian SMPL $\beta$ distributions (right), as proposed by \cite{sengupta2021probabilisticposeshape}, cannot model local shape uncertainty arising from ambiguous inputs, since $\beta$s control global deformations over the whole body surface. Global shape uncertainty is less useful for downstream probabilistic combination as it does not specify which local body parts are uncertain.}
    \label{fig:pred_meas_dist_vis}
\end{figure}

3D human shape and pose estimation from RGB images is a challenging computer vision problem, with direct applications in virtual retail, virtual reality and computer animation. Several deep-learning-based approaches to this task yield impressive human pose estimates \cite{hmrKanazawa17, kolotouros2019cmr, kolotouros2019spin, zhang2019danet, georgakis2020hkmr, Guler_2019_CVPR_holopose, Moon_2020_ECCV_I2L-MeshNet, Choi_2020_ECCV_Pose2Mesh}. However, body shape estimates tend to be inaccurate or inconsistent for subjects in-the-wild. Recently, \cite{sengupta2021probabilisticposeshape, sengupta2021hierprobposeshape} attempt to predict accurate and consistent body shapes from multiple images of a subject, without assuming a fixed body pose or background and lighting conditions. This involves (i) predicting independent Gaussian distributions (i.e. with diagonal covariance matrices) over SMPL \cite{SMPL:2015} shape parameter vectors conditioned on the input images and (ii) probabilistically combining the shape distributions predicted from each image, to yield a final consistent shape estimate. However, independent Gaussian distributions over SMPL shape parameters are unable to quantify uncertainty in \textit{local} body parts, since SMPL shape parameters (i.e. coefficients of a PCA shape space) control shape deformation over the \textit{global} body surface. Given multiple images of a subject, meaningful probabilistic shape combination benefits from local shape uncertainty estimation, where part-specific uncertainty arises from variation in camera viewpoints and body poses within the images, as well as occlusion (see Figures \ref{fig:pred_meas_dist_vis} and \ref{fig:pred_meas_dist_vis_synth}).

\begin{figure}[t!]
    \centering
    \includegraphics[width=\textwidth]{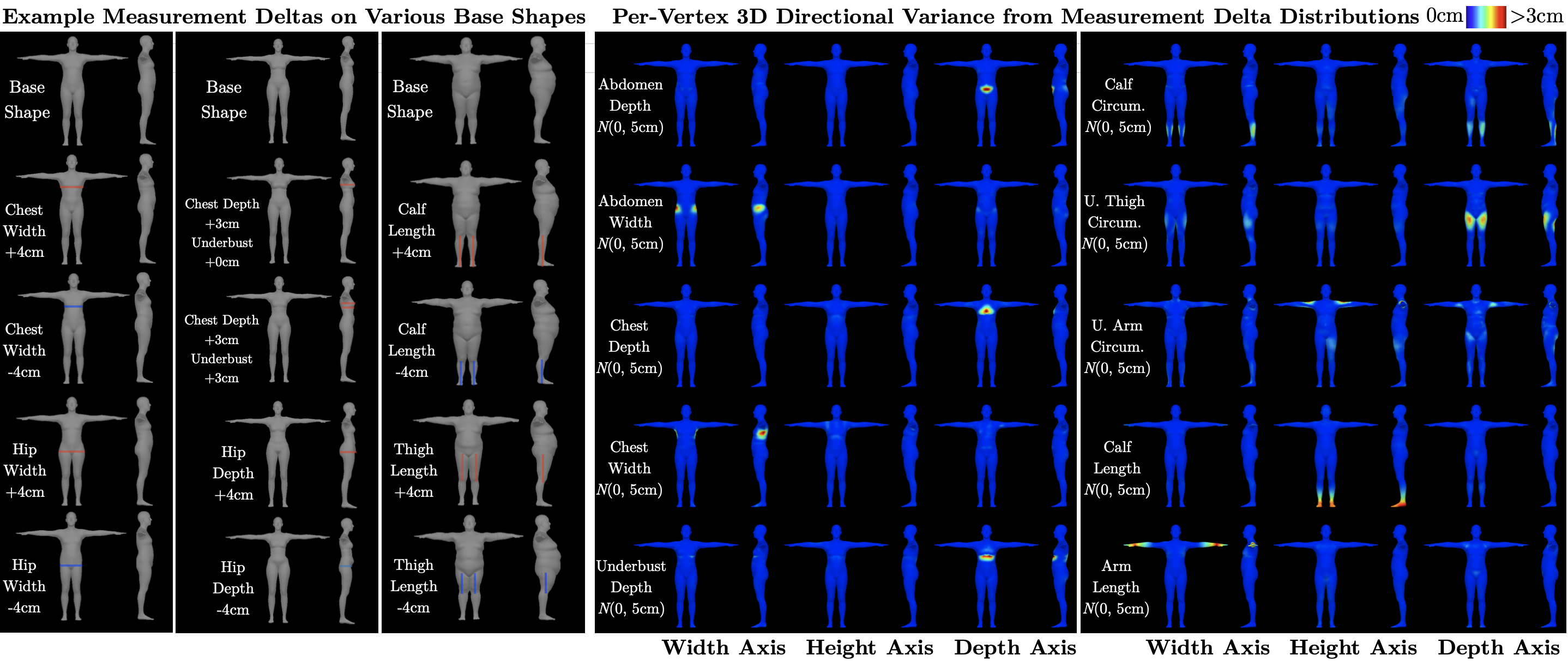}
    \caption{Left: Effect of various input measurement offsets applied to 3 different base body shapes. Measurement offsets are mapped to SMPL $\beta$ offsets using the proposed measurements-to-$\beta$s linear regressor. Right: Transformation of Gaussian distributions over various measurement offsets to Gaussians over 3D vertex locations. Visualisation of 3D vertex variances along the width ($x$), height ($y$) and depth ($z$) axes aligned with the front-facing human body. Note that semantic measurement offsets result in local shape deformations, and distributions over these offsets result in localised vertex variance, representing uncertainty in each vertex's 3D location in the T-pose.}
    \label{fig:meas_def_dist}
\end{figure}

To this end, we extend \cite{sengupta2021probabilisticposeshape} by predicting distributions over local semantic body shape \textit{measurements} (\eg chest width, arm length, calf circumference, etc), conditioned on an input image. This necessitates learning a mapping from semantic body measurements to SMPL shape coefficients ($\beta$s), which enables local, human-interpretable control of SMPL body shapes. Independent Gaussian distributions defined over measurements translate to localised uncertainty over SMPL T-pose vertices (as shown in Figure \ref{fig:meas_def_dist}), in contrast with independent Gaussian distributions over SMPL $\beta$s. Furthermore, we define the mapping from measurements to SMPL $\beta$s to be a linear regression. Thus, a Gaussian distribution over measurements can be analytically transformed into a distribution over SMPL $\beta$s, and subsequently 3D vertex locations, using simple linear transformations (see Equation \ref{eqn:lineargaussianproject}).

\begin{figure}[t!]
    \centering
    \includegraphics[width=\textwidth]{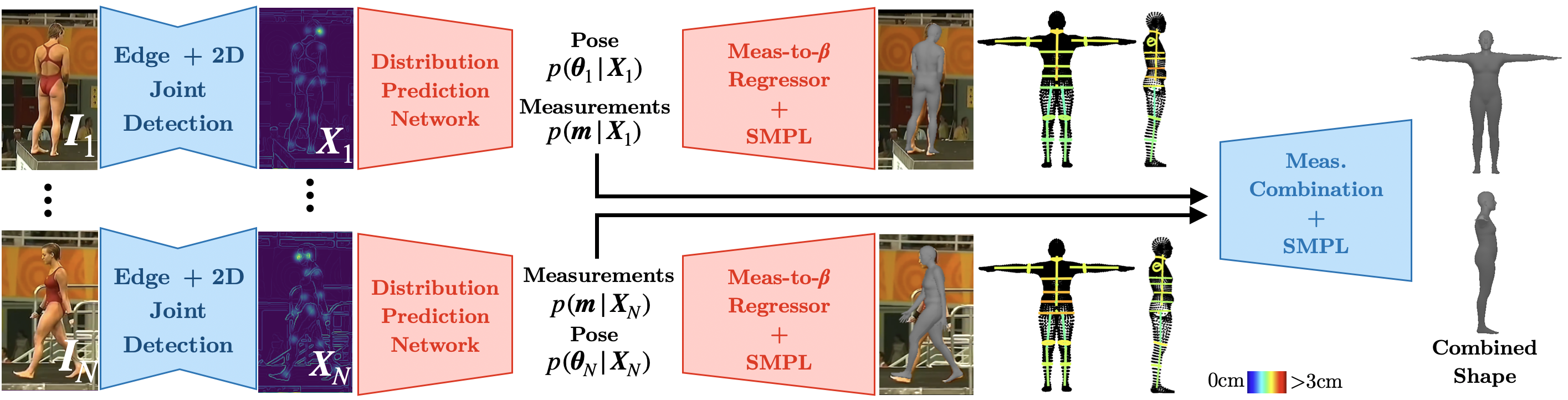}
    \vspace{-0.3cm}
    \caption{Our three-stage approach for body measurement and pose estimation from a set of images. Each image is converted into an edge and joint heatmap proxy representation, which is passed through a distribution prediction network that yields distributions over body measurements and pose conditioned on the input images. Individual measurement distributions are probabilistically combined into a final measurement estimate, which can be mapped to SMPL shape coefficients using the proposed measurements-to-$\beta$s linear regressor.}
    \label{fig:method}
\end{figure}

Having learned a linear mapping from measurements to SMPL $\beta$s, our pipeline for 3D multi-image body shape and pose estimation consists of 3 stages (see Figure \ref{fig:method}). First, we compute proxy representations using an off-the-shelf 2D keypoint detector \cite{he2017maskrcnn, wu2019detectron2} and Canny edge detection \cite{canny1986edge}. This decreases the domain gap between synthetic training data and real test data \cite{sengupta2021hierprobposeshape}. Second, a deep neural network predicts means and variances of Gaussian distributions over SMPL pose parameters and body measurements, conditioned on the input proxy representations. Third, body measurements from each image are probabilistically combined \cite{sengupta2021probabilisticposeshape} to give a final measurements estimate, which is converted into a full body shape estimate using our measurements-to-$\beta$s regressor and the SMPL function \cite{SMPL:2015}. Probabilistic combination intuitively amounts to uncertainty-weighted averaging (Equation \ref{eqn:probcomb}) - since our measurement distributions are able to better capture \textit{local} shape uncertainty than independent Gaussians over SMPL $\beta$s \cite{sengupta2021probabilisticposeshape}, we obtain improved body shape estimation accuracy. This is quantitatively corroborated by shape metrics on the SSP-3D dataset \cite{STRAPS2020BMVC}, as well two private datasets of tape-measured humans, in an A-pose and in varying poses.

\section{Related Work}

This section reviews recent approaches to 3D human shape and pose estimation from images.

\noindent \textbf{Monocular shape and pose estimators} may be classified as optimisation-based or learning-based. Optimisation-based approaches fit a parametric 3D body model \cite{SMPL:2015, SMPL-X:2019, Joo_2018_CVPR_total_capture, osman202star} to 2D observations, such as 2D keypoints \cite{Bogo:ECCV:2016, SMPL-X:2019, Lassner:UP:2017}, silhouettes \cite{Lassner:UP:2017} or part segmentations \cite{Zanfir_2018_CVPR} by minimising a suitable error function. They do not require expensive 3D-labelled training, but are sensitive to poor initialisations and inaccurate 2D observations. Learning-based approaches can be further split into model-free or model-based. Model-free methods train deep networks to directly predict human body meshes \cite{kolotouros2019cmr, Choi_2020_ECCV_Pose2Mesh, Moon_2020_ECCV_I2L-MeshNet, Zeng_2020_CVPR_mesh_dense}, voxel occupancies \cite{varol18_bodynet} or implicit surface representations \cite{saito2019pifu, saito2020pifuhd} given an input image. Model-based methods \cite{hmrKanazawa17, georgakis2020hkmr, zhang2019danet, Guler_2019_CVPR_holopose, omran2018nbf, kolotouros2019spin, tan2017, pavlakos2018humanshape, biggs2020multibodies} regress 3D body model parameters \cite{SMPL:2015, SMPL-X:2019, osman202star}, which give a low-dimensional representation of a 3D human body. Learning-based methods yield impressive 3D pose estimates in-the-wild, but shape predictions are often inaccurate, due to the lack of shape diversity in training datasets. Some recent works improve shape estimates using synthetic training data \cite{STRAPS2020BMVC, sengupta2021probabilisticposeshape, sengupta2021hierprobposeshape, smith20193dfromsilhouettes}, which we adopt in our method.

\noindent \textbf{Multi-image shape and pose estimators} leverage the extra shape information present in videos \cite{kocabas2019vibe, humanMotionKanazawa19, tung2017selfsupmocap, pavlakos2019texturepose, Arnab_CVPR_2019, alldieck2017optical}, as well as multi-view images \cite{liang2019samv, smith20193dfromsilhouettes} of a subject in a fixed pose captured from multiple camera angles. In contrast, \cite{sengupta2021probabilisticposeshape} propose to estimate body shape from a set of \textit{unconstrained} images of a subject, by probabilistically-combining distributions over SMPL \cite{SMPL:2015} shape parameters. We extend this work by predicting distributions over local body measurements instead of global shape parameters, and demonstrate that this improves shape estimation accuracy from sets of unconstrained images.

\vspace{-0.5cm}
\section{Method}

This section provides a brief overview of SMPL \cite{SMPL:2015}, introduces our measurements-to-$\beta$s linear regressor and presents our three-stage pipeline for probabilistic human pose and body measurement estimation from multiple images of a subject.

\noindent \textbf{SMPL} \cite{SMPL:2015} is a parametric 3D human body model. It provides a differentiable function that maps pose parameters $\boldsymbol{\theta}$, shape parameters $\boldsymbol{\beta}$ and global body rotation $\boldsymbol{\gamma}$ to a 3D vertex mesh $\boldsymbol{V} \in \mathbb{R}^{6890 \times 3}$. $\boldsymbol{\theta}$ represents 3D joint rotations, relative to each joint's parent in the kinematic tree, in axis-angle form (i.e. $\boldsymbol{\theta} \in \mathbb{R}^{69}$ for 23 SMPL joints). Similarly, $\boldsymbol{\gamma} \in \mathbb{R}^3$ represents root joint rotation (i.e. global body orientation) in axis-angle form. The shape parameter vector $\boldsymbol{\beta} \in \mathbb{R}^{|\boldsymbol{\beta}|}$ consists of coefficients quantifying the contribution of PCA shape-space basis vectors,  $\{\boldsymbol{S}_i\}_{i=1}^{|\boldsymbol{\beta}|}$ where $\boldsymbol{S}_i \in \mathbb{R}^{6890 \times 3}$, to the identity-dependent body shape. Specifically, shape-space basis vectors represent deformations from a template mesh $\boldsymbol{T} \in \mathbb{R}^{6890 \times 3}$ over the full body surface. The identity-dependent (i.e. T-pose) 3D vertex mesh is then given by
\begin{equation}
\boldsymbol{\tilde{V}} = \sum_{i=1}^{|\boldsymbol{\beta}|} \beta_i \boldsymbol{S}_i + \boldsymbol{T} = \text{vec}^{-1} (\mathcal{S} \boldsymbol{\beta} + \boldsymbol{t})
\label{eqn:smpl_shape}
\end{equation}
where $\boldsymbol{t} = \text{vec}(\boldsymbol{T}) \in \mathbb{R}^{20670} $ and $\mathcal{S}= [\text{vec}(\boldsymbol{S}_1), ...,\text{vec}(\boldsymbol{S}_{|\boldsymbol{\beta}|})]  \in \mathbb{R}^{20670 \times |\boldsymbol{\beta}|}$ represent shape-space bases and template vertices flattened with the $\text{vec}()$ operation. $\text{vec}^{-1}()$ denotes the inverse, converting a vector back into a matrix containing 3D vertices.

\noindent \textbf{Measurements-to-$\beta$s Linear Regressor.} We learn a simple linear regressor from 23 body measurements to SMPL shape coefficients. Please refer to the supplementary material for a list of measurements used and details regarding the definition of measurements over a SMPL T-pose body, which is abstracted here as an operation $\boldsymbol{m} = \text{measure}(\boldsymbol{\beta})$ that outputs body measurements $\boldsymbol{m} \in \mathbb{R}^{23}$ given shape coefficients $\boldsymbol{\beta} \in \mathbb{R}^{|\boldsymbol{\beta}|}$. We aim to obtain a mapping from measurement \textit{offsets} $\boldsymbol{\Delta m}$ to shape coefficient \textit{offsets} $\boldsymbol{\Delta \beta}$, such that
\begin{equation}
\boldsymbol{\Delta \beta}^T = \boldsymbol{\Delta m}^T \boldsymbol{W}
\label{eqn:meas2betas}
\end{equation}
where $\boldsymbol{W} \in \mathbb{R}^{23 \times |\boldsymbol{\beta}|}$ is the weight matrix of the linear regressor. Then, each specific measurement of a given base body shape $\boldsymbol{\beta}$ (with measurements $\boldsymbol{m}$) can be offset by (i) setting the corresponding element of $\boldsymbol{\Delta m}$ to the desired value, (ii) obtaining $\boldsymbol{\Delta \beta}$ using Equation \ref{eqn:meas2betas}, and (iii) adding the shape offset to the base shape to yield a new body $\boldsymbol{\beta} + \boldsymbol{\Delta \beta}$ with measurements $\boldsymbol{m} + \boldsymbol{\Delta m}$. Several measurement offsets on varying base body shapes are visualised in Figure \ref{fig:meas_def_dist} (left). Note that the mean SMPL body is given by $\boldsymbol{\bar{\beta}} = \boldsymbol{0}$ with measurements $\boldsymbol{\bar{m}} = \text{measure}(\boldsymbol{\bar{\beta}})$. Thus, if the base body shape is assumed to be the mean SMPL body $\boldsymbol{\bar{\beta}} = \boldsymbol{0}$, coefficient offsets $\boldsymbol{\Delta \beta}$ are equivalent to the new shape coefficients themselves.

To learn the weight matrix $\boldsymbol{W}$, we first randomly sample a range of SMPL shape coefficients and stack them into a matrix $\boldsymbol{B} \in \mathbb{R}^{L \times |\boldsymbol{\beta}|}$, with $L=10^6$ samples. Corresponding measurements are obtained as $\boldsymbol{M} = \text{measure}(\boldsymbol{B}) \in \mathbb{R}^{L \times 23}$. SMPL mean shape and measurements are subtracted to give  $\boldsymbol{\Delta B} =\boldsymbol{B} - \boldsymbol{\bar{\beta}} $ and  $\boldsymbol{\Delta M} = \boldsymbol{M} - \boldsymbol{\bar{m}}$. Then, $\boldsymbol{W}$, such that $\boldsymbol{\Delta B} = \boldsymbol{\Delta M} \boldsymbol{W}$, is estimated in a least squares sense using the pseudo-inverse $\boldsymbol{W} = (\boldsymbol{\Delta M}^T \boldsymbol{\Delta M})^{-1}\boldsymbol{\Delta M}^T \boldsymbol{\Delta B}$. 

An independent Gaussian distribution over measurement offsets, $\mathcal{N}(\boldsymbol{\mu_{\Delta m}}, \text{diag}(\boldsymbol{\sigma^2_{\Delta m}}))$, can be transformed to a Gaussian distribution over shape coefficients, $\mathcal{N}(\boldsymbol{\mu_{\beta}}, \boldsymbol{\Sigma_{\beta}})$, and then over T-pose vertices $\mathcal{N}(\boldsymbol{\mu_{\tilde{V}}}, \boldsymbol{\Sigma_{\tilde{V}}})$ using linear transformations of Gaussians (note that the SMPL mean shape $\boldsymbol{\bar{\beta}} = \boldsymbol{0}$ is assumed as the base body shape):
\begin{equation}
\begin{aligned}
\boldsymbol{\mu_{\beta}} = \boldsymbol{W}^T \boldsymbol{\mu_{\Delta m}}, \;\;
\boldsymbol{\Sigma_{\beta}} = \boldsymbol{W}^T \text{diag}(\boldsymbol{\sigma^2_{\Delta m}}) \boldsymbol{W}, \;\;\;\;
\boldsymbol{\mu_{\tilde{V}}} = \mathcal{S}\boldsymbol{\mu_{\beta}}  + \boldsymbol{t}, \;\; \boldsymbol{\Sigma_{\tilde{V}}} = \mathcal{S} \boldsymbol{\Sigma_{\beta}} \mathcal{S}^T
\end{aligned}
\label{eqn:lineargaussianproject}
\end{equation}
which follows from Equations \ref{eqn:smpl_shape} and \ref{eqn:meas2betas}. The diagonal terms of the covariance matrix $\boldsymbol{\Sigma_{\tilde{V}}}$ quantify the variance (i.e. uncertainty) in the 3D locations of T-pose vertices in the $x$, $y$ and $z$ directions (i.e. width, height and depth axis). Figure \ref{fig:meas_def_dist} (right) visualises these directional variances given different input measurement offset distributions.

\noindent \textbf{Proxy representation computation.} Given $N$ RGB images $\{\boldsymbol{I}_n\}_{n=1}^N$ of a subject, we first compute edge-images and 2D joint heatmaps (see Figure \ref{fig:method}), using Canny edge detection \cite{canny1986edge} and Detectron2 \cite{wu2019detectron2}. The edge-image and joint heatmaps of $\boldsymbol{I}_n \in \mathbb{R}^{H \times W \times 3}$ are stacked to form a proxy representation $\boldsymbol{X}_n\ \in \mathbb{R}^{H \times W \times (J+1)}$ (for $J$ joints). We use this proxy representation as our input, instead of the RGB image, to decrease the domain gap between synthetic training images \cite{STRAPS2020BMVC, sengupta2021probabilisticposeshape, sengupta2021hierprobposeshape} and real test images.

\noindent \textbf{Body measurements and pose distribution prediction.} Next, we follow \cite{sengupta2021probabilisticposeshape} and pass each $\boldsymbol{X}_n$ into a distribution prediction neural network (as shown in Figure \ref{fig:method}). However, instead of predicting a distribution over SMPL shape coefficients, our network outputs the means and variances of independent Gaussian distributions over \textit{measurement offsets} $\boldsymbol{\Delta m}$ (from the mean SMPL body measurements $\boldsymbol{\bar{m}}$), as well as pose parameters $\boldsymbol{\theta}$, conditioned on the inputs:
\begin{equation}
p(\boldsymbol{\theta}_n | \boldsymbol{X}_n) = \mathcal{N}(\boldsymbol{\mu_{\theta}}({\boldsymbol{X}_n}),  \boldsymbol{\Sigma_{\theta}}({\boldsymbol{X}_n})), \;\;
p(\boldsymbol{\Delta m} | \boldsymbol{X}_n) = \mathcal{N}(\boldsymbol{\mu_{\Delta m}}({\boldsymbol{X}_n}), \boldsymbol{\Sigma_{\Delta m}}({\boldsymbol{X}_n}))
\label{eqn:preddist}
\end{equation}
where $\boldsymbol{\Sigma_{\theta}}({\boldsymbol{X}_n}) = \text{diag}(\boldsymbol{\sigma^2_{\theta}}({\boldsymbol{X}_n}))$ and $\boldsymbol{\Sigma_{\Delta m}}({\boldsymbol{X}_n}) = \text{diag}(\boldsymbol{\sigma^2_{\Delta m}}({\boldsymbol{X}_n}))$. Specifically, $\boldsymbol{\sigma^2_{\theta}}({\boldsymbol{X}_n})$ and $\boldsymbol{\sigma^2_{\Delta m}}({\boldsymbol{X}_n})$ estimate the heteroscedastic aleatoric uncertainty \cite{derkiureghian2009aleatoric_epistemic, kendall2017whatuncertainties} in pose and measurement predictions, arising from ambiguities in the inputs due to varying camera views and poses (resulting in self-occlusion), or occluding objects. Furthermore, our network also outputs per-image deterministic estimates of weak-perspective camera parameters $\{\boldsymbol{c}_n\}_{n=1}^N$, representing scale and $xy$ translation, and global body rotations $\{\boldsymbol{\gamma}_n\}_{n=1}^N$. 

As an aside, our measurements-to-$\beta$s regressor, in theory, can be subsumed into the SMPL $\beta$ distribution network of \cite{sengupta2021probabilisticposeshape}. However, in practice, \textit{independent} Gaussian distributions (with diagonal covariances) over SMPL $\beta$s cannot model local shape uncertainty. We would need to predict Gaussian distributions with full $|\boldsymbol{\beta}|\times |\boldsymbol{\beta}|$ positive semi-definite covariance matrices (see Equation 3). This is difficult compared to (i) learning the measurements-to-$\beta$s regressor separately, and (ii) predicting per-measurement variances $\boldsymbol{\sigma^2_{\Delta m}}({\boldsymbol{X}_n}) \in \mathbb{R}^{23}$.

\noindent \textbf{Multi-image measurement combination.} Finally, we implement a similar probabilistic combination operation to \cite{sengupta2021probabilisticposeshape}, that combines the shape distributions from the individual images into a final, consistent body shape. However, instead of combining predicted distributions over SMPL shape coefficients, our combination is done in the body measurement space using the predicted measurement distributions:
\vspace{-0.2cm}
\begin{equation}
\begin{aligned}
&p(\boldsymbol{\Delta m} | \{\boldsymbol{X}_n\}_{n=1}^N) 
\propto \prod_{n=1}^N p(\boldsymbol{\Delta m} | \boldsymbol{X}_n)
\propto \mathcal{N}(\boldsymbol{\Delta m}; \boldsymbol{\mu_\text{comb}}, \boldsymbol{\Sigma_\text{comb}})\\
&\boldsymbol{\Sigma_\text{comb}} = \bigg(\sum_{n=1}^N \boldsymbol{\Sigma^{-1}_{\Delta m}} ({\boldsymbol{X}_n})\bigg)^{-1}, \;\;
\boldsymbol{\mu_\text{comb}} = \boldsymbol{\Sigma_\text{comb}}\bigg(\sum_{n=1}^N \boldsymbol{\Sigma_{\Delta m}}^{-1}({\boldsymbol{X}_n})\boldsymbol{\mu_{\Delta m}}({\boldsymbol{X}_n})\bigg).
\end{aligned}
\label{eqn:probcomb}
\vspace{-0.2cm}
\end{equation}
We observe that combining measurement distributions instead of shape coefficient distributions results in improved shape estimation accuracy (see Section \ref{sec:experimental_results}), since distributions over measurements are able to predict local shape uncertainty due to varying camera views, poses and occlusions, unlike independent Gaussian distributions over global shape coefficients (see Figures \ref{fig:pred_meas_dist_vis} and \ref{fig:pred_meas_dist_vis_synth}). Please refer to \cite{sengupta2021probabilisticposeshape} for more details on probabilistic shape combination.

At any stage of the inference pipeline, predicted measurement distributions or final combined measurement estimates can be easily converted into SMPL shape coefficient distributions/estimates using Equations \ref{eqn:meas2betas} and \ref{eqn:lineargaussianproject}.

\noindent \textbf{Loss functions.} At test-time, our measurement and pose prediction pipeline deals with sets of input images. However, training occurs using a dataset of single-image input-label pairs, denoted by $\{\boldsymbol{X}_k, \{\boldsymbol{\theta}_k, \boldsymbol{\Delta m}_k, \boldsymbol{\gamma}_k\}\}_{k=1}^K$, with $K$ i.i.d training samples. Note that measurement offset labels $\boldsymbol{\Delta m}_k$ represent offsets from the mean SMPL body measurements $\boldsymbol{\bar{m}}$. 

We train the distribution prediction network with a negative log-likelihood loss $\mathcal{L}_{\text{NLL}} = - \sum_{k=1}^K \bigg( \log p(\boldsymbol{\theta}_k |\boldsymbol{X}_k) + \log p(\boldsymbol{\Delta m}_k |\boldsymbol{X}_k) \bigg)$.
We also apply the same 2D joints samples loss proposed in \cite{sengupta2021probabilisticposeshape}, as well as a mean-squared-error loss over global body rotation matrices.

\begin{table}[t!]
\renewcommand{\tabcolsep}{4pt}
\footnotesize
    \centering
     \begin{tabular}{c||ccccccc||cc}
     \textbf{Num. $\beta$s} & \multicolumn{7}{c||}{\textbf{Local Offset Evaluation}} & \multicolumn{2}{c}{ \textbf{Reconstruction Eval.}}\\
        \textbf{Used} & \textbf{Input Meas. $\Delta$} & \multicolumn{6}{c||}{\textbf{Output Meas. $\Delta$}} & \textbf{Meas. MAE} & \textbf{PVE-T}\\
        & & Ch. W. & Ch. D. & St. W. & St. D. & Ca. C. & Ca. L.\\
        \hline
        \multirow{3}{1em}{10} & Ch. W. +50 & \underline{+27.3} & +5.0 & +10.1 & -4.1 & +6.1 & -1.1 & \multirow{3}{1em}{\textbf{0.9}} & \multirow{3}{1em}{\textbf{1.9}}\\
        & St. D. +50 & -2.7 & +7.8 & +11.2 & \underline{+29.9} & +6.0 & +5.8\\
        & Ca. L. +50 & -3.1 & +0.8 & +5.3 & +5.2 & -2.5 & +\underline{27.8} \\
        \hline
         \multirow{3}{1em}{70} & Ch. W. +50 & \underline{\textbf{+51.1}} & +0.5 & +0.5 & +0.0 & +0.3 & +0.0 & \multirow{3}{1em}{3.9} & \multirow{3}{1em}{15.4}\\
        & St. D. +50 & -0.3 & -0.5 & +0.0 & \underline{\textbf{+49.8}} & -3.6 & +0.0\\
        & Ca. L. +50 & +0.0 & +0.2 & +0.3 & +1.9 & +0.2 & \underline{\textbf{+50.1}}\\
        \hline
         \multirow{3}{1em}{90} & Ch. W. +50 & \underline{+51.6} & +0.4 & +0.0 & +0.5 & +0.4 & +0.0 & \multirow{3}{1em}{6.2} & \multirow{3}{1em}{23.9}\\
        & St. D. +50 & +0.0 & +0.1 & +0.0 & \underline{+54.3} & -0.6 & +0.0\\
        & Ca. L. +50 & +0.1 & -0.2 & -0.4 & -1.0 & +1.7 & \underline{+50.3}\\
    \end{tabular}
    \caption{Local controllability and reconstruction ability of our measurements-to-$\beta$s regressor, using different numbers of SMPL $\beta$s. Local Offset Evaluation involves passing an input offset of +50mm for chest width, stomach depth and calf length in turn through the linear regressor, and computing the corresponding output measurement offsets, thereby quantifying the local controllability of our approach. Reconstruction Evaluation quantifies how well an SMPL body (represented by T-pose vertices) can be reconstructed from just its corresponding measurements, in terms of measurement error (Meas. MAE) and per-vertex error (PVE-T). These are computed by sampling 100,000 random input SMPL bodies, passing their measurements through the linear regressor and comparing the output SMPL bodies with the inputs. All numbers in mm. Abbreviations: Ch. = Chest, St. = Stomach, Ca. = Calf, W. = Width, D. = Depth, C. = Circumference, L. = Length. }
    \label{tab:meas2betas_eval}
    \vspace{-0.2cm}
\end{table}
\vspace{-0.3cm}

\section{Implementation Details}
\noindent \textbf{Network architecture.} Our distribution prediction network consists of a ResNet-18 \cite{He2015resnet} convolutional encoder, followed by a 3 layer fully-connected network with 512 neurons in the two hidden layers and ELU activations \cite{clevert2016elu}, and 190 output neurons. Output variances are forced to be positive using an exponential activation function.\\
\noindent \textbf{Synthetic training.} We adopt the training frameworks presented in \cite{STRAPS2020BMVC, smith20193dfromsilhouettes, pavlakos2018humanshape, sengupta2021probabilisticposeshape}, which entail on-the-fly generation of synthetic training inputs and corresponding SMPL body shape and pose labels during training. In short, for each training iteration, ground-truth body shapes are randomly sampled from a Gaussian distribution over the SMPL shape space, while ground-truth poses are obtained from the training sets of UP-3D \cite{Lassner:UP:2017}, 3DPW \cite{vonMarcard2018} and H3.6M \cite{h36m_pami}. These are rendered into synthetic input proxy representations using the SMPL function, a light-weight renderer \cite{ravi2020pytorch3d} and Canny edge detection \cite{canny1986edge}. Synthetic inputs are augmented using various occlusion and corruption transforms. Our method differs from past work in 2 main ways: (i) our measurements-to-$\beta$s regressor allows us to randomly sample measurement offsets to further augment the random SMPL shape samples, and (ii) we use body measurement labels to train our network using $\mathcal{L}_{\text{NLL}}$, and thus need to compute ground-truth measurements from the sampled ground-truth shape coefficients.\\
\noindent \textbf{Training details.} We use Adam \cite{kingma2014adam} with a learning rate of 0.0001, batch size of 80 and train for 150 epochs, which takes 2 days on a 2080Ti GPU.\\
\noindent \textbf{Evaluation datasets.} SSP-3D \cite{STRAPS2020BMVC} is used to evaluate body shape prediction accuracy in the wild. We report per-vertex Euclidean error in the T-pose after scale correction (PVE-T-SC) \cite{STRAPS2020BMVC} and mean absolute measurement error after scale correction (Meas. MAE-SC), both in mm. In addition, we evaluate on two private datasets of tape-measured humans: ``A-Pose Subjects'' consists of front and side views of 8 subjects in an A-pose and ``Varying-Pose Subjects'' consists of 27 images of 4 subjects in a range of poses and camera views. We report chest, stomach and hip circumference measurement errors. The test set of 3DPW \cite{vonMarcard2018} is used to evaluate body pose accuracy, using mean-per-joint-position-error after scale correction (MPJPE-SC) and after Procrustes analysis (MPJPE-PA). Finally, we utilise a synthetic evaluation dataset for our ablation studies, which consists of 1000 synthetic subjects with randomly sampled body measurements. Each subject is posed using 4 SMPL poses sampled from Human3.6M \cite{h36m_pami} and global body rotations are set to face forward, backwards, left and right. Synthetic evaluation inputs are generated in the same way as our training inputs.

Please refer to the supplementary material for further details regarding synthetic data generation, training hyperparameters and example test images from the private shape evaluation datasets with tape-measured humans and the synthetic ablation dataset.

\begin{table}[t!]
\scriptsize
\renewcommand{\tabcolsep}{5pt}
    \centering
    \begin{tabular}{l||ccc||ccc|ccc||cc }
        \multirow{3}{1em}{\textbf{Method}} & \multicolumn{3}{c||}{\textbf{Synthetic}} & \multicolumn{6}{c||}{\textbf{SSP-3D}} & \multicolumn{2}{c}{\textbf{3DPW}}\\
        & \multicolumn{3}{c||}{\textbf{Meas. MAE-SC}} & \multicolumn{3}{c|}{\textbf{Meas. MAE-SC}} & \multicolumn{3}{c||}{\textbf{PVE-T-SC}} & 
        \textbf{MPJPE-SC} & \textbf{MPJPE-PA}\\
        & SI & NA & PC & SI & NA & PC & SI & NA & PC\\
        \hline
        10 $\beta$s Net & 23.3 & 18.5  & 18.3  & 23.4 & 20.2 & 20.2 & 13.6 & 13.0 & 12.8 & 87.3 & 60.3\\
        70 $\beta$s Net & 23.2 & 18.6 & 18.2 & 23.7 & 20.1 & 20.0 & 13.6 & 12.9 & 12.9 & 92.9 & 63.8\\
        Measure Net (Ours) & 21.6 & 17.8 & \textbf{16.4} & 22.8 & 19.9 & \textbf{19.5} & 13.7 & 12.8 & \textbf{12.4} & 88.3 & 61.6\\
        \hline
        GraphCMR \cite{kolotouros2019cmr}& - & - & - & 47.2 & 47.0 & - & 19.5 & 19.3 & - & 102.0 & 70.2\\
        SPIN \cite{kolotouros2019spin}& - & - & - & 49.8 & 49.7 & - & 22.2 & 21.9 & - & 89.4 & 59.2\\
        DaNet \cite{zhang2019danet}& - & - & - & 49.9 & 49.7 & - & 22.1 & 22.1 & - & \textbf{82.4} & \textbf{54.8}\\
        STRAPS \cite{STRAPS2020BMVC}& - & - & - & 24.5 & 21.0 & - & 15.9 & 14.4 & - & 99.0 & 66.8\\
        Sengupta \etal \cite{sengupta2021probabilisticposeshape} & - & - & - & 24.4 & 20.6 & 20.4 & 15.2 & 13.6 & 13.3 & 90.9 & 61.0\\
        VIBE* \cite{kocabas2019vibe} & - & - & - & - & 50.1 & - & - & 24.1 & - & - & \textbf{51.9}\\
    \end{tabular}
    \caption{Single-image (SI) and multi-image (NA/PC) body shape evaluation on synthetic ablation data and SSP-3D \cite{STRAPS2020BMVC}, and single-image pose evaluation on 3DPW \cite{vonMarcard2018}. Multi-image shape evaluation compares naive-averaging (NA) of shapes predicted from individual inputs against probabilistic shape combination (PC) (i.e. uncertainty-weighted averaging). The top half compares SMPL shape $\beta$ distribution predictors against our measurement distribution predictor (both trained on the same synthetic data). The bottom half presents metrics from competing approaches. Note that our 10 $\beta$s Net is equivalent to \cite{sengupta2021probabilisticposeshape}, except we use improved edge-based training inputs \cite{sengupta2021hierprobposeshape}. All numbers in mm. *VIBE \cite{kocabas2019vibe} uses video inputs.}
    \vspace{-0.1in}
    \label{tab:ssp3d_3dpw_eval}
\end{table}

\section {Experimental Results}
\label{sec:experimental_results}

\begin{figure}[t!]
    \centering
    \includegraphics[width=\textwidth]{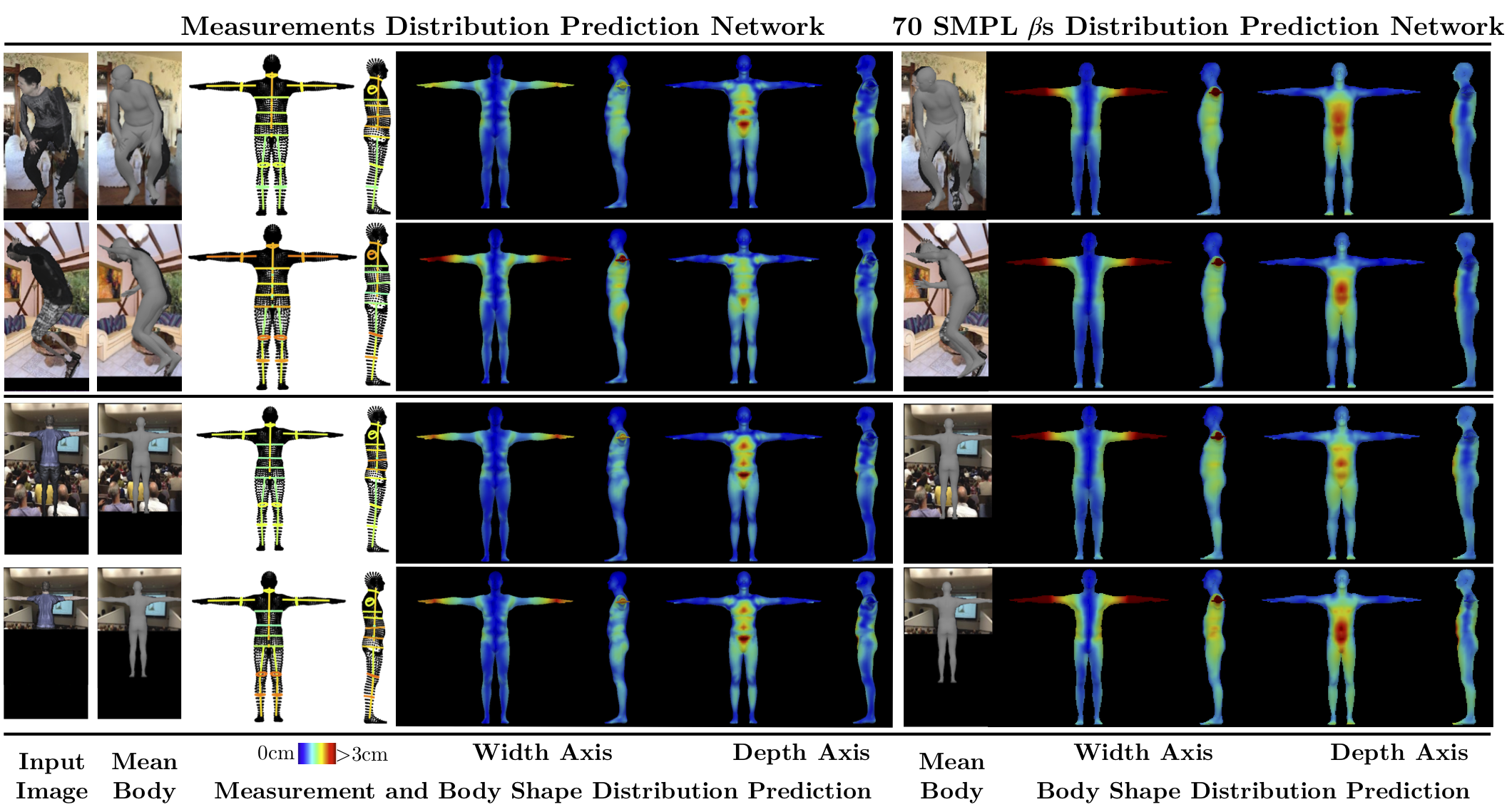}
    \caption{Comparing independent Gaussian measurement distributions and SMPL $\beta$ distributions on synthetic images. Similar to Figure \ref{fig:pred_meas_dist_vis}, measurement distributions (left) capture \textit{local} shape uncertainty. For example, a front-on camera viewpoint (row 1) results in larger predicted depth measurement uncertainties (column 4) compared to a side-on viewpoint (row 2). Bent limbs result in higher limb length uncertainties (see arm lengths in row 1 vs row 2). Furthermore, when body parts are locally occluded, measurements \textit{specific to the occluded part} have larger predicted uncertainties (see rows 3-4, columns 3-8). In contrast, independent Gaussian $\beta$ distributions \cite{sengupta2021probabilisticposeshape} (right) cannot model local shape uncertainty. For example, a locally occluded input results in an undesired increase of \textit{global} shape uncertainty over the whole body surface (see rows 3-4, columns 10-13), which is less useful for downstream probabilistic combination as it does not specify which body parts are uncertain.}
    \vspace{-0.1in}
    \label{fig:pred_meas_dist_vis_synth}
\end{figure}

This section discusses our ablation studies on the measurements-to-$\beta$s linear regressor, compares measurement versus shape coefficient distribution prediction and evaluates the performance of our method on real datasets against state-of-the-art approaches.

\noindent \textbf{Measurements-to-$\beta$s linear regressor.} Table \ref{tab:meas2betas_eval} investigates the proposed measurements-to-$\beta$s regressor using $|\boldsymbol{\beta}| =$ 10, 70 and 90 SMPL shape coefficients. From the local offset analysis (Table \ref{tab:meas2betas_eval}, left), it is clear that 10 shape PCA coefficients are not expressive enough to locally control body shape, since an input offset for one measurement (\eg +50mm chest width in row 1) results in significant output offsets for several other measurements. Increasing the number of PCA shape coefficients used, from 10 to 70, greatly improves the local controllability of the model, but further increasing to 90 does not provide much additional benefit. Qualitative examples of local offsets are given in Figure \ref{fig:meas_def_dist}.

Conversely to the local offset analysis, using a larger number of shape coefficients increases reconstruction error (Table \ref{tab:meas2betas_eval}, right). While the greater expressiveness of more shape coefficients is beneficial for local offsets, it means that measurements alone do not contain enough information to reconstruct full body T-pose meshes. As a compromise between local controllability and reconstruction error, we use 70 SMPL shape coefficients.

\noindent \textbf{Comparison between $\beta$ distributions and measurement distributions.} Table \ref{tab:ssp3d_3dpw_eval} (top) compares our proposed measurement distribution prediction network against SMPL $\beta$ distribution predictors using 10 and 70 $\beta$s. Probabilistic combination (i.e. uncertainty-weighted averaging) using $\beta$ distributions only results in marginal improvements over naive-averaging of $\beta$s, since the predicted $\beta$ distributions are unable to quantify local uncertainty. In contrast,  probabilistic \textit{measurement} combination yields a significant improvement over naive-averaging of measurements, on both synthetic data and SSP-3D, since measurement distributions capture local shape uncertainty due to varying poses/camera angles (which cause self-occlusion), as well as occluding objects (see Figures \ref{fig:pred_meas_dist_vis} and \ref{fig:pred_meas_dist_vis_synth}). Table \ref{tab:tapemeaseval} (top) further exhibits the benefits of probabilistic measurement combination on both A-pose humans and humans in varying poses. Note that combining $\beta$ distributions (rows 1-2) can even result in \textit{worse} measurement errors than naive-averaging for varying-pose subjects (showcasing challenging body poses), while measurement combination always improves errors. 

Moreover,  Table \ref{tab:ssp3d_3dpw_eval} shows a reduction in pose estimation accuracy for $\beta$ distribution predictors when the number of $\beta$s used is increased from 10 to 70. We hypothesise that it is challenging for the network to learn to estimate distributions over $7\times$ more shape parameters, and pose accuracy suffers as a result. In contrast, predicting distributions over 23 local body measurements allows us to benefit from the increased expressiveness of 70 shape coefficients, without compromising pose.

\begin{table}[t!]
\renewcommand{\tabcolsep}{3.5pt}
\scriptsize
    \centering
    \begin{tabular}{l||ccc|ccc|ccc||ccc|ccc|ccc}
        \multirow{3}{1em}{\textbf{Method}} & \multicolumn{9}{c||}{\textbf{A-Pose Subjects}} & \multicolumn{9}{c}{\textbf{Varying-Pose Subjects}}\\
        & \multicolumn{3}{c|}{\textbf{Chest}} & \multicolumn{3}{c|}{\textbf{Stomach }} & \multicolumn{3}{c||}{\textbf{Hip }}& \multicolumn{3}{c|}{\textbf{Chest}} & \multicolumn{3}{c|}{\textbf{Stomach }} & \multicolumn{3}{c}{\textbf{Hip }}\\
        & SI & NA & PC & SI & NA & PC & SI & NA & PC & SI & NA & PC & SI & NA & PC & SI & NA & PC\\
        \hline
        10 $\beta$s Net & 63 & 54 & 52 & 61 & 54 & 52 & 54 & 38 & 35 & 83 & 72 & 76 & 47 & 30 & 32 & 42 & 27 & 27\\
        70 $\beta$s Net & 61 & 44 & 39 & 61 & 47 & 45 & 52 & 32 & 25 & 60 & 56 & 58 & 39 & 32 & 30 & 33 & 27 & 25\\
        Measure Net (Ours) & 52 & 37 & \textbf{33} & 37 & 32 & \textbf{29} & 37 & 29 & \textbf{23} & 63 & 53 & \textbf{52} & 43 & 29 & \textbf{27} & 37 & 23 & \textbf{17}\\
        \hline
        SPIN \cite{kolotouros2019spin} & 130 & 127 & - & 117 & 114 & - & 125 & 124 & - & 57 & 56 & - & 60 & 60 & - & 74 & 73 &- \\
        STRAPS \cite{STRAPS2020BMVC} & 82 & 80 & - & 81 & 81 & - & 84 & 83 & - & 67 & 67 & - & 54 & 53 & - & 59 & 55 & -\\
        Sengupta \etal \cite{sengupta2021probabilisticposeshape} & 65 & 53 & 51 & 61 & 54 & 52 & 53 & 49 & 42 & 78 & 68 & 67 & 49 & 41 & 43 & 42 & 33 & 35\\
    \end{tabular}
    \caption{Single-image (SI) and multi-image (NA/PC) body shape evaluation on two datasets of tape-measured humans, containing subjects in an A-pose and subjects in varying poses respectively. Multi-image shape evaluation compares naive-averaging (NA) of shapes predicted from individual inputs and probabilistic body shape combination (PC) (i.e. uncertainty-weighted averaging). The top half compares SMPL shape $\beta$ distribution predictors against our proposed measurement distribution predictor. The bottom half presents metrics from competing state-of-the-art approaches. All numbers are circumference errors in mm. Note that our 10 $\beta$s Net is equivalent to \cite{sengupta2021probabilisticposeshape}, except we use improved edge-based training inputs \cite{sengupta2021hierprobposeshape}.}
    \vspace{-0.1in}
    \label{tab:tapemeaseval}
\end{table}

\noindent \textbf{Comparison with the state-of-the-art.} Table 2 (bottom) presents shape and pose metrics from several approaches evaluated on SSP-3D and 3DPW. Our probabilistic measurement combination approach yields the best shape metrics on SSP-3D. In terms of pose metrics, it is competitive with approaches that do not any require 3D-labelled training images \cite{STRAPS2020BMVC, sengupta2021probabilisticposeshape}. Table 3 (bottom) also shows that measurement combination outperforms all other approaches in terms of measurement errors on both A-pose and varying-pose subjects.

\vspace{-0.3cm}
\section{Conclusion}
In this work, we propose a locally controllable shape model by learning a linear mapping from semantic body measurements to SMPL \cite{SMPL:2015} shape $\beta$s. This is motivated by the observation that distributions over SMPL shape $\beta$s are unable to meaningfully capture shape uncertainty associated with \textit{locally}-occluded body parts, since the SMPL shape space represents \textit{global} deformations over the whole body surface.
Our measurements-to-$\beta$s regressor allows us to predict distributions over body measurements conditioned on input images. We demonstrate the value of the proposed procedure when predicting a body shape estimate from a set of images of a subject, where we achieve state-of-the-art identity-dependent body shape estimation accuracy on the SSP-3D~\cite{sengupta2021probabilisticposeshape} dataset and a private dataset of tape-measured humans, using probabilistic measurement combination.

 \begin{raggedright}
    {\LARGE \bfseries\sffamily\textcolor{bmv@sectioncolor}{Supplementary Material: Probabilistic Estimation of 3D Human Shape and Pose with a Semantic Local Parametric Model}\par}
  \end{raggedright}
  \vskip\baselineskip
  \hrule
  \vskip\baselineskip

This document provides additional material supplementing the main manuscript. Section \ref{supmat_sec:body_meas_def} gives definitions of width, depth, circumference and length measurements over the SMPL \cite{SMPL:2015} body surface. Section \ref{supmat_sec:local_control} qualitatively corroborates the local controllability experiments presented in the main manuscript. Section \ref{supmat_sec:qualitative_results} provides qualitative results comparing our measurement distribution prediction network against previously-proposed \cite{sengupta2021probabilisticposeshape} SMPL shape coefficient ($\beta$) distribution predictors, using input images from our (i) synthetic evaluation dataset, (ii) SSP-3D \cite{STRAPS2020BMVC}, and (iii) two private datasets of tape-measured humans, which were named ``A-Pose Subjects'' and ``Varying-Pose Subjects'' in the main manuscript. Section \ref{supmat_sec:synthetic_training} contains details regarding synthetic data generation and examples of synthetic training images and synthetic evaluation images used in the ablation studies presented in the main manuscript.

\vspace{-0.3cm}
\section{Body Measurement Definitions}
\label{supmat_sec:body_meas_def}

\begin{figure}[t!]
    \centering
    \includegraphics[width=0.7\textwidth]{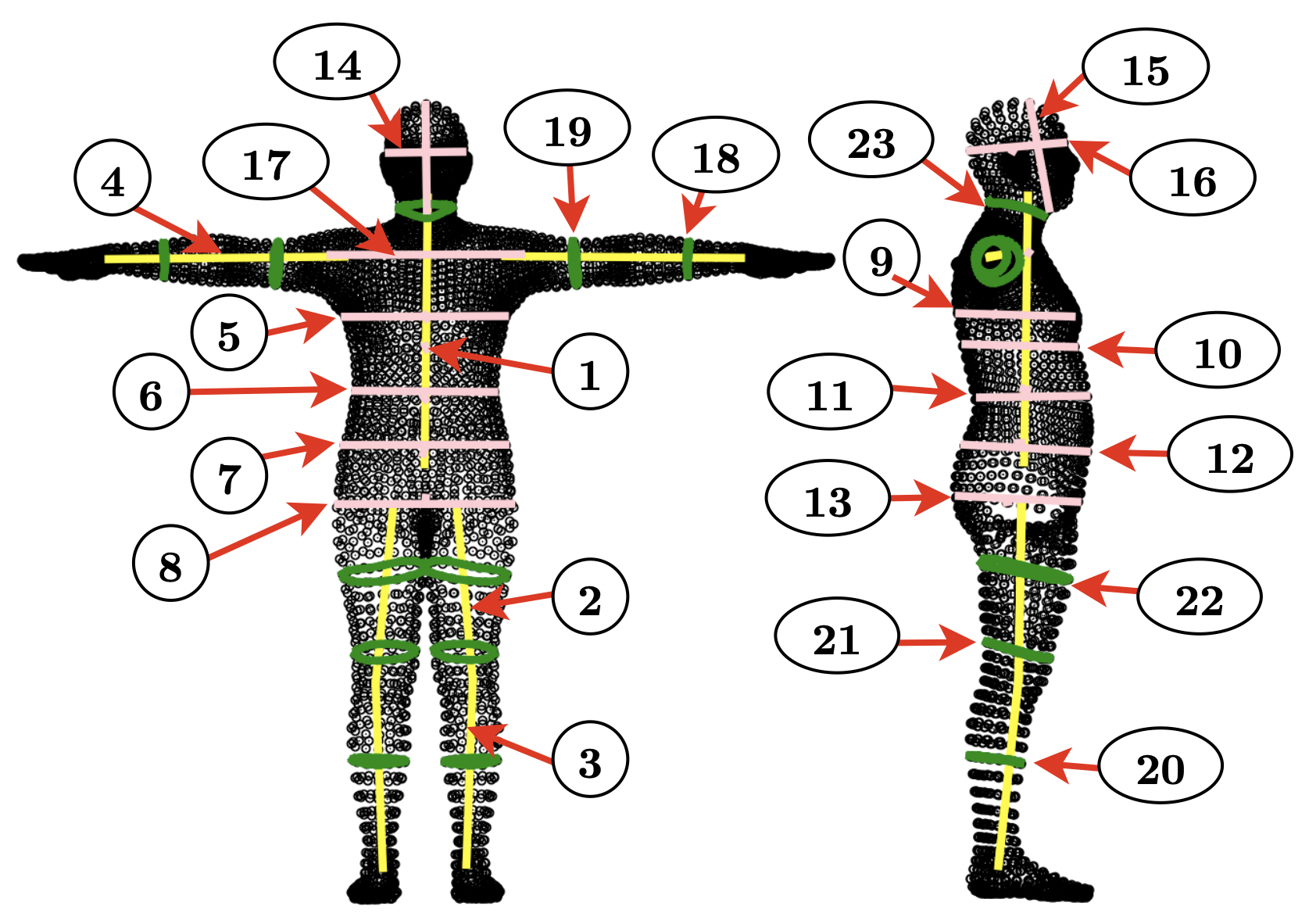}
    \caption{Front- and side-view visualisation of 23 measurement definitions over the SMPL \cite{SMPL:2015} body surface. Please refer to Figure \ref{fig:meas_def_table} for the semantic meaning of each measurement and the specific SMPL vertex/joint IDs used to define them. Colour key: Yellow = ``Length'' measurements defined using SMPL T-pose joints, Pink = ``Width'' / ``Depth'' measurements defined using SMPL T-pose vertices, Green = ``Circumference'' measurements defined using SMPL T-pose vertices.}
    \label{fig:meas_def_vis}
\end{figure}

\begin{figure}[h!]
    \centering
    \includegraphics[width=\textwidth]{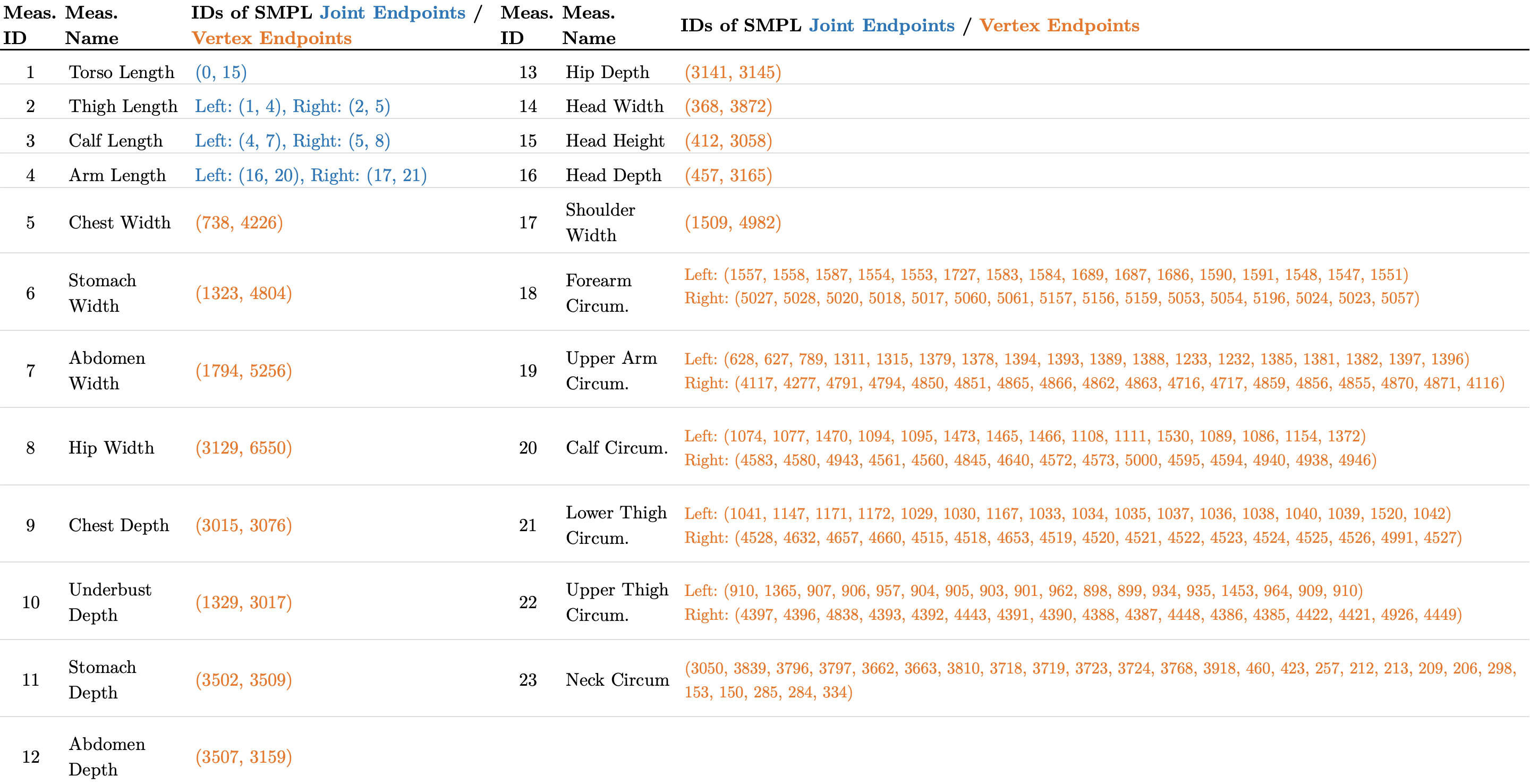}
    \caption{Semantic meaning of each measurement visualised in Figure \ref{fig:meas_def_vis}, along with SMPL joint/vertex IDs used to define them. Joint/vertex IDs correspond to endpoints for ``Width'', ``Depth'' and ``Length'' measurements, and waypoints for ``Circum.'' measurements. These specific 23 measurements were chosen to sufficiently constrain the body surface, such that a full T-pose body mesh can be recovered from just 23 measurements.}
    \label{fig:meas_def_table}
\end{figure}

In the main manuscript, obtaining measurements from an SMPL T-pose body was abstracted away as an operation $\boldsymbol{m} = \text{measure}(\boldsymbol{\beta})$. In this section, Figures \ref{fig:meas_def_vis} and \ref{fig:meas_def_table} give definitions of each of the 23 body measurements, in terms of the SMPL T-pose joint/vertex IDs used as endpoints (for widths, depths and lengths) or waypoints (for circumferences). Given the joint/vertex IDs, measurement values are obtained by simply computing the 3D Euclidean distance between the corresponding endpoints/waypoints for a T-pose body. For circumferences, the Euclidean distance between waypoints are summed along the circumference. Note that a T-pose SMPL body only depends on given shape coefficients $\boldsymbol{\beta}$ (see Equation 1 in the main manuscript). Thus, the operation $\boldsymbol{m} = \text{measure}(\boldsymbol{\beta})$ involves (i) generating T-pose joints and vertices from the input $\boldsymbol{\beta}$, (ii) gathering measurement endpoints/waypoints using the joint and vertex IDs given in Figure \ref{fig:meas_def_table} and (iii) computing Euclidean distances between endpoints/waypoints and summing if needed.

Figures \ref{fig:meas_def_vis} and \ref{fig:meas_def_table} visualise and define separate left/right limb measurements. However, locally controlling left/right measurements independently is challenging with the SMPL body model. The SMPL shape space is learnt using PCA applied to human body scans from CAESAR \cite{CAESAR2002} and the majority of human bodies exhibit strong left/right symmetry, both in CAESAR and in the general population. Thus, we convert the separate left/right limb measurements defined in Figures \ref{fig:meas_def_vis} and \ref{fig:meas_def_table} into single limb measurements by taking the mean of the left and right sides.

\section{Local Controllability}
\label{supmat_sec:local_control}

\begin{figure}[t!]
    \centering
    \includegraphics[width=\textwidth]{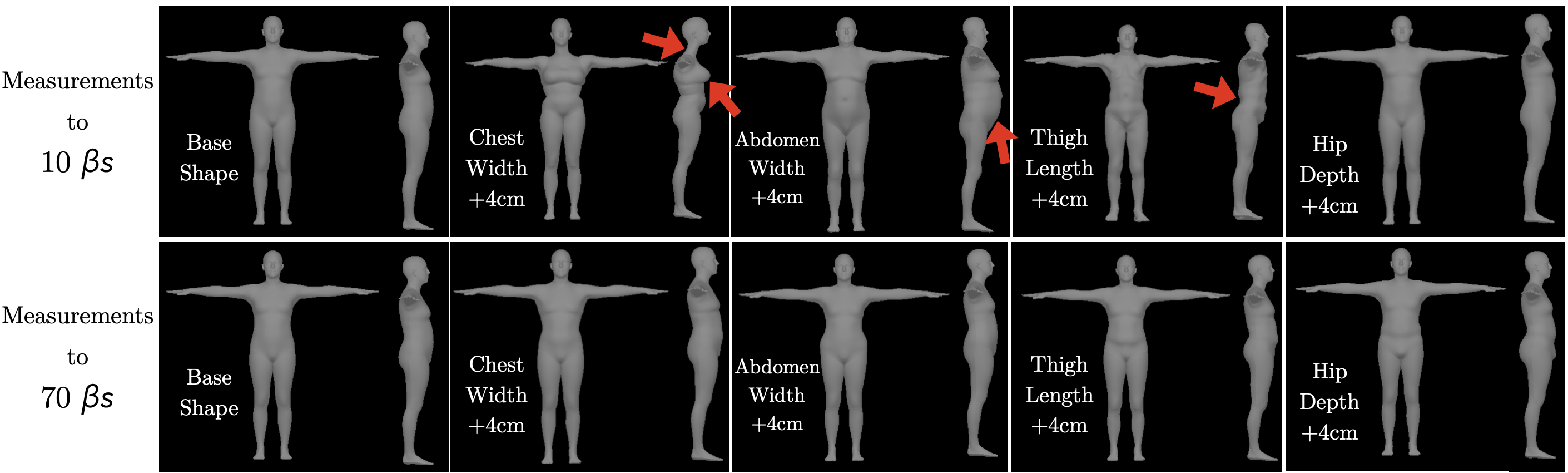}
    \caption{Visualisation of the effect of +4cm input measurement offsets applied to a base body shape. Measurement offsets are mapped to SMPL $\beta$ offsets using the proposed measurements-to-$\beta$s linear regressor, using 10 SMPL $\beta$s (top) and 70 SMPL $\beta$s (bottom). Regressing only 10 $\beta$s may result in (i) non-local output offsets and unrealistic output body shapes (top row, highlighted by red arrows) or (ii) zero output offsets and unchanged output body shapes (top row, hip depth in column 5 ). Regressing 70 $\beta$s yields \textit{realistic and local} body shape offsets, which match the desired inputs.}
    \label{fig:10vs70betas_vis}
\end{figure}

The main manuscript analyses the local controllability of the proposed measurements-to-$\beta$s regressor. In particular, we quantitatively show that regressing from body measurements to 10 SMPL shape coefficients ($\beta$s) results in poor controllability, wherein an input offset of +5cm applied to a specific measurement results in large undesired output offsets to several other measurements. This is because the 10-dimensional SMPL shape space is not expressive enough to allow for fine-grained local control of body shape. A significant quantitative improvement in local controllability is observed when the number of SMPL $\beta$s is increased from 10 to 70. Figure \ref{fig:10vs70betas_vis} demonstrates this qualitatively, by visualising the effect of input measurement offsets on SMPL bodies when regressing 10 $\beta$s and 70 $\beta$s. Using only 10 $\beta$s may result in either (i) non-local output offsets and unrealistic output body shapes (Figure \ref{fig:10vs70betas_vis}, top row, highlighted by red arrows) or (ii) zero output offsets and unchanged output body shapes (Figure \ref{fig:10vs70betas_vis}, top row, column 5). On the other hand, the measurement-to-70-$\beta$s regressor yields \textit{realistic and local} body shapes offsets, which match the desired inputs.

\begin{figure}[t!]
    \centering
    \includegraphics[width=\textwidth]{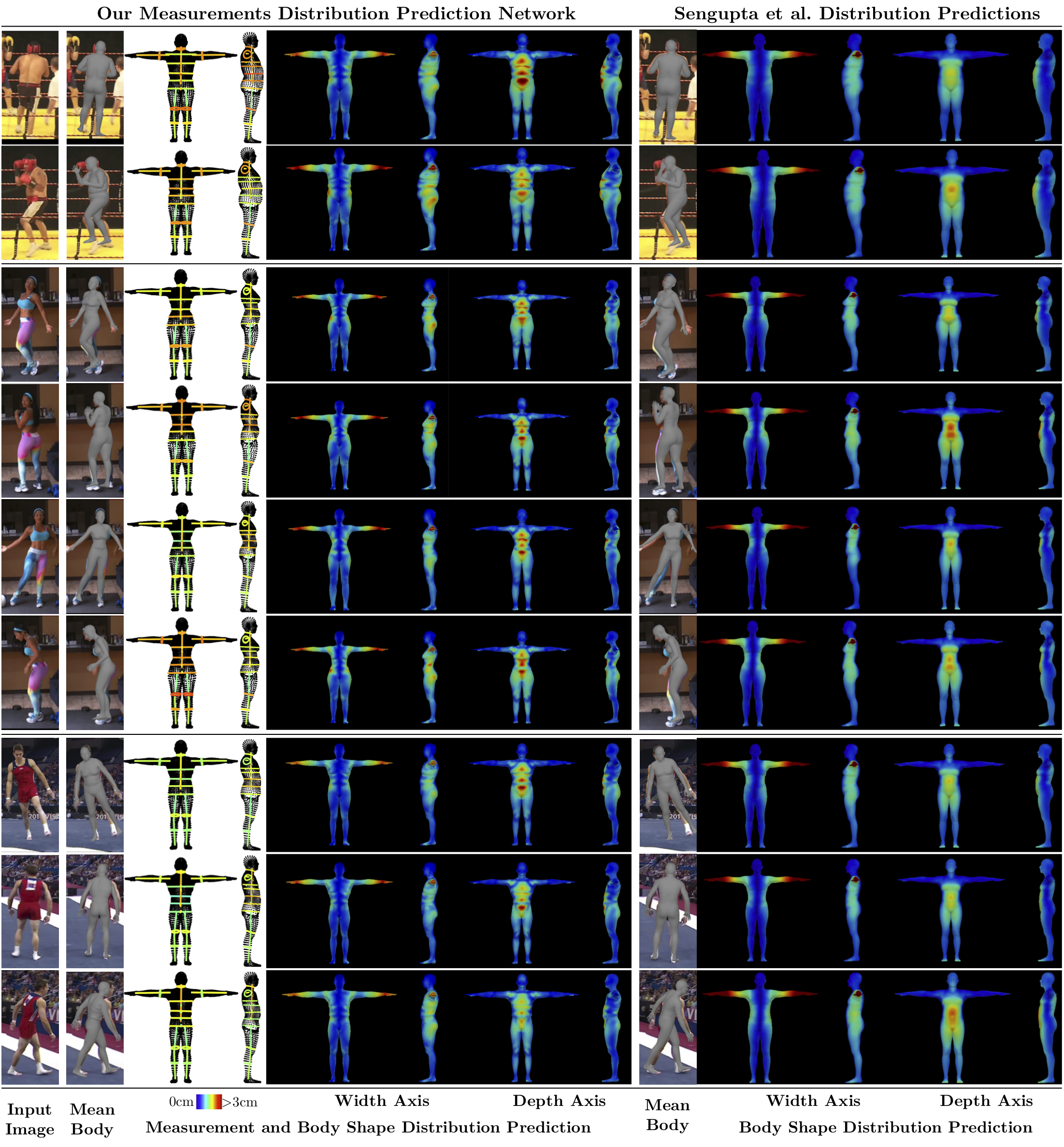}
    \caption{Comparison between our predicted measurement distributions and SMPL $\beta$ distributions from Sengupta \etal \cite{sengupta2021probabilisticposeshape} on images from SSP-3D \cite{STRAPS2020BMVC}. Note that \cite{sengupta2021probabilisticposeshape} is the previous state-of-the-art approach in terms of body shape metrics on SSP-3D. Similar to Figure 3 in the main manuscript, this figure demonstrates that Gaussian measurement distributions exhibit meaningful \textit{local} shape uncertainty arising from varying camera angles, challenging poses and self-occlusions. In contrast, independent Gaussian SMPL $\beta$ distributions from \cite{sengupta2021probabilisticposeshape} cannot model such local shape uncertainty, since $\beta$s control global deformations over the whole body surface. Instead, predicted shape uncertainty increases globally over the whole body when the input contains a challenging pose or self-occlusion (compare the bottom 2 rows, columns 10-13). Such global uncertainty is less useful for downstream tasks as it does not specify which body-parts have high prediction uncertainty, only that the network is uncertain as a whole. Thus, probabilistically combining measurement distributions yields better shape metrics than \cite{sengupta2021probabilisticposeshape}, as shown in Tables 2 and 3 in the main manuscript.}
    \label{fig:ssp3d_sota_compare}
\end{figure}

\section{Qualitative Results}
\label{supmat_sec:qualitative_results}

\begin{figure}[h!]
    \centering
    \includegraphics[width=\textwidth]{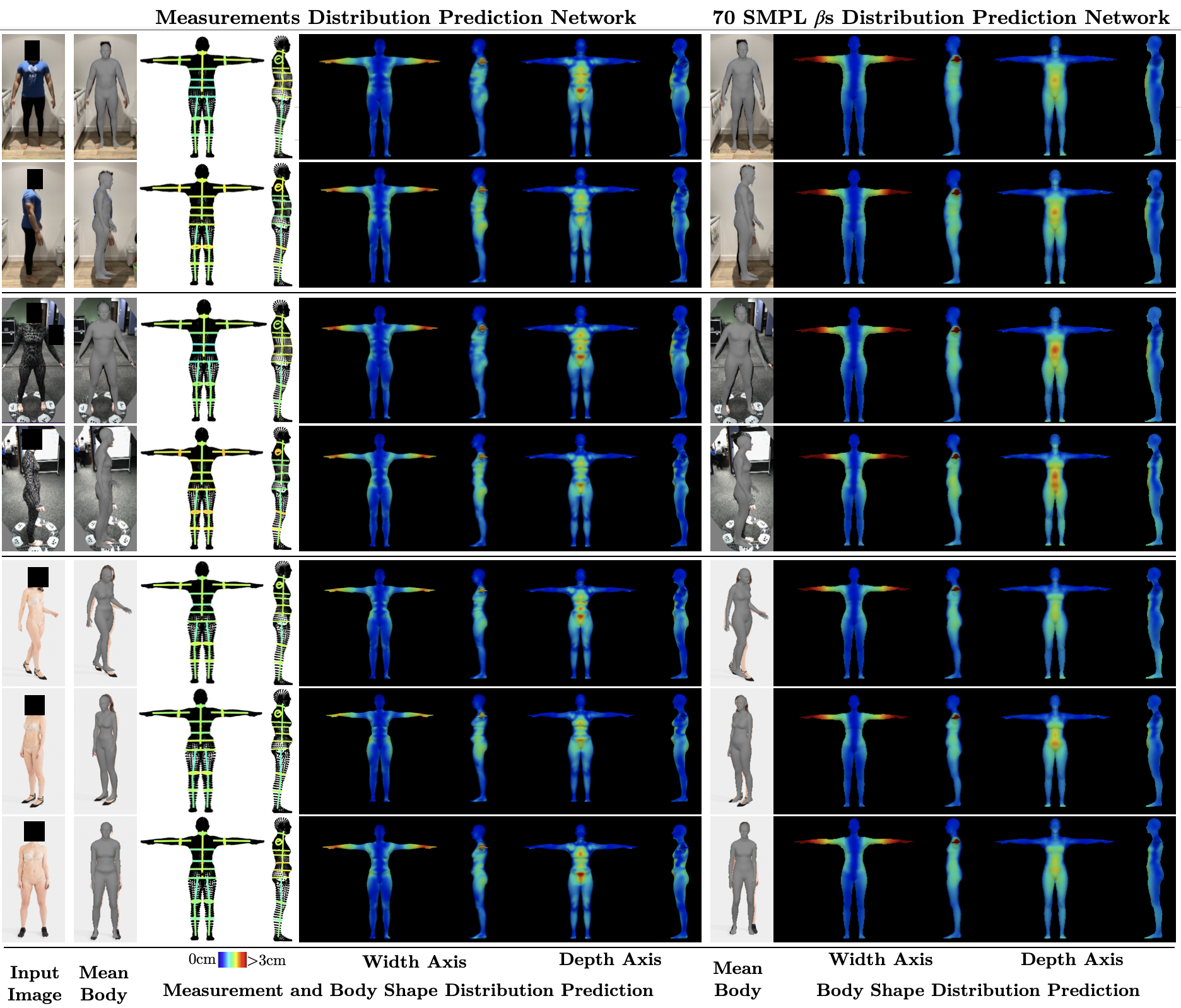}
    \caption{Comparison between measurement distribution predictions and SMPL 70 $\beta$ distribution predictions on images from our two private datasets of tape-measured humans, ``A-Pose Subjects'' (rows 1-4) and ``Varying-Pose Subjects'' (rows 5-7). This figure further corroborates that predicting measurement distributions leads to meaningful local shape uncertainty when given images with different global body orientations, i.e. front-facing images result in lower predicted uncertainties for width measurements (columns 3, 5, 6), while side-facing images result in lower uncertainties for  depth measurements (columns 4, 7, 8). On the other hand, uncertainty predictions from the SMPL 70 $\beta$ distribution network cannot model such local ambiguities in the input image due to variations in camera viewpoint.}
    \label{fig:metail_pred}
\end{figure}

\begin{figure}[h!]
    \centering
    \includegraphics[width=\textwidth]{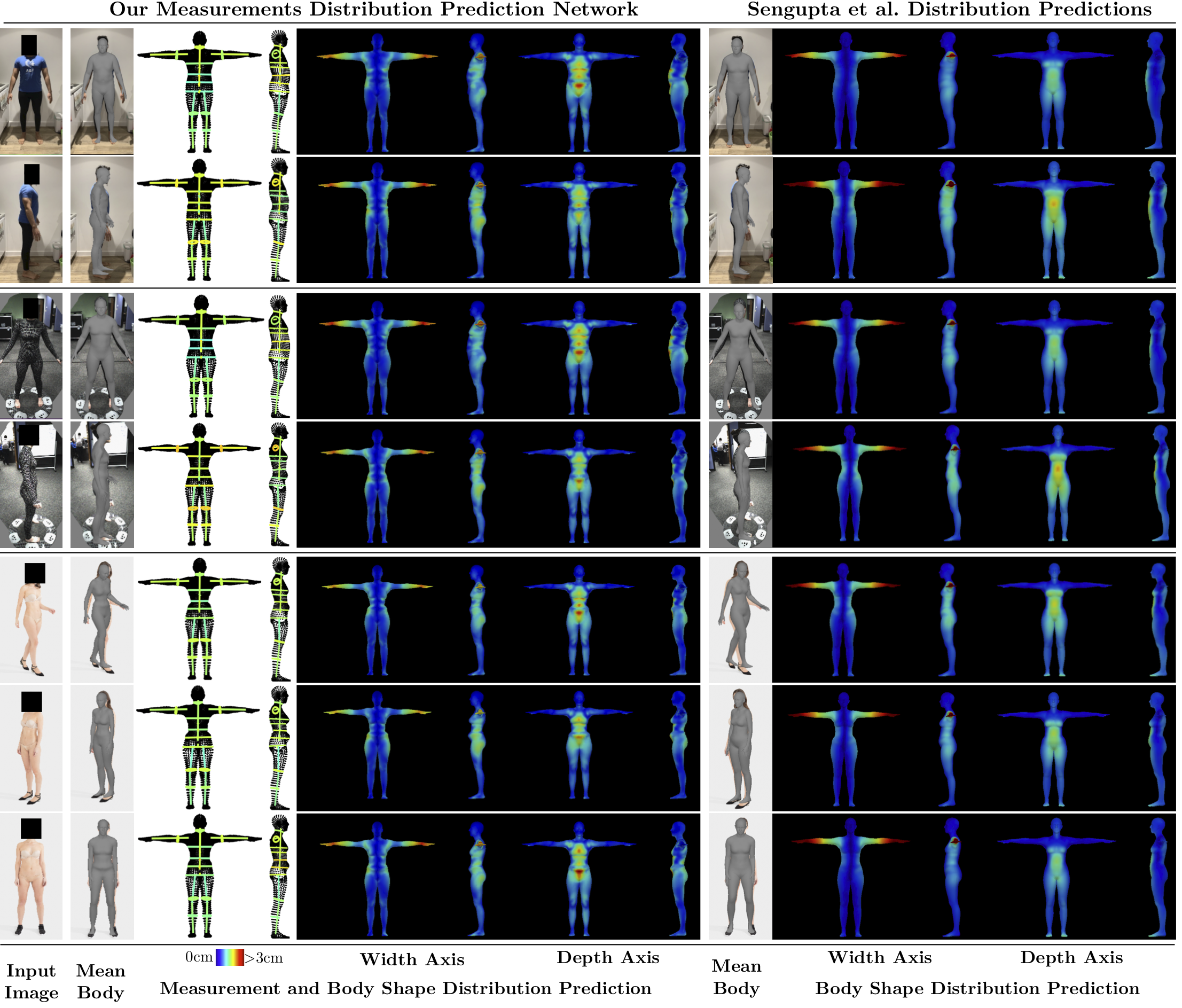}
    \caption{Comparison between our predicted Gaussian measurement distributions and SMPL $\beta$ distributions from Sengupta \etal \cite{sengupta2021probabilisticposeshape} on images from our two private datasets of tape-measured humans, ``A-Pose Subjects'' (rows 1-4) and ``Varying-Pose Subjects'' (rows 5-7). Similar to Figure \ref{fig:metail_pred}, measurement distribution prediction results in intuitive local shape uncertainty when given images with different global body orientations, while predictions from the SMPL $\beta$ distribution network of \cite{sengupta2021probabilisticposeshape} do not appear to be related to the shape information present in the input (as dictated by the subject's global body orientation, pose or occlusions).}
    \label{fig:metail_sota_compare}
\end{figure}

Figures \ref{fig:ssp3d_sota_compare}, \ref{fig:metail_pred} and \ref{fig:metail_sota_compare} compare results from our proposed measurement distribution predictor and SMPL $\beta$ distribution predictors, both using 70 $\beta$s (i.e. the number of shape coefficients output by our measurements-to-$\beta$s regressor), as well as 10 $\beta$s as proposed by Sengupta \etal \cite{sengupta2021probabilisticposeshape}. We conclude that predicting Gaussian distributions over semantic body measurements allows for meaningful predictions of \textit{local} aleatoric \cite{kendall2017whatuncertainties, derkiureghian2009aleatoric_epistemic} shape uncertainty that models ambiguities in the input images related to the subject's pose, camera viewpoint and occlusion, which is not possible with independent Gaussian distributions over \textit{global} SMPL $\beta$s. Improved local shape uncertainty quantification yields better body shape estimates after probabilistic combination, as demonstrated in the main manuscript.

\begin{figure}[h!]
    \centering
    \includegraphics[width=\textwidth]{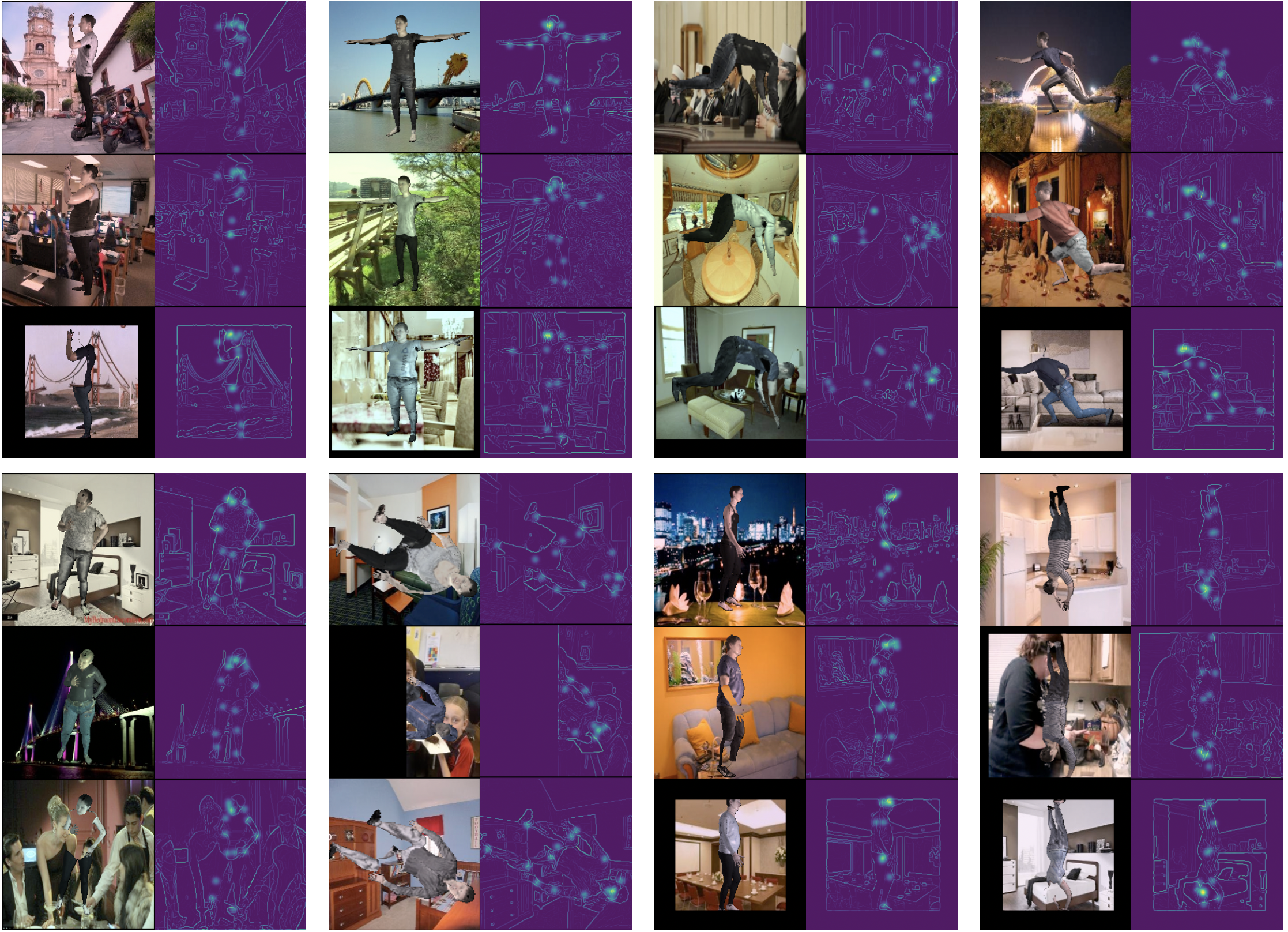}
    \caption{Examples of synthetic RGB training images and corresponding edge-image + 2D joint heatmap proxy representations. Images within each group of 3 use the same pose (selected from common 3D SMPL pose datasets \cite{h36m_pami, vonMarcard2018, Lassner:UP:2017}) but different random body shapes, clothing textures, backgrounds and lighting parameters. The synthetic RGB images are computationally cheap and far from photorealistic - however, edge-filtering significantly reduces the synthetic-to-real domain gap.}
    \label{fig:synthetic_training_examples}
\end{figure}

\section{Synthetic Data Generation}
\label{supmat_sec:synthetic_training}

Following \cite{STRAPS2020BMVC, sengupta2021probabilisticposeshape, sengupta2021hierprobposeshape, smith20193dfromsilhouettes}, we adopt a synthetic training framework as a means of overcoming the lack of body shape diversity in common datasets for 3D pose and shape estimation from images \cite{vonMarcard2018, h36m_pami, Lassner:UP:2017}. In particular, we use a edge-and-joint-heatmap proxy representation \cite{sengupta2021hierprobposeshape}, to bridge the domain gap between low-fidelity synthetic training inputs and real test inputs.

Examples of synthetic RGB images, and corresponding edge-image + 2D joint heatmap proxy representations, are given in Figure \ref{fig:synthetic_training_examples}. They are generated on-the-fly during training by sampling a random SMPL shape, SMPL pose, clothing texture and background image for each training iteration, and rendering using a light-weight renderer \cite{ravi2020pytorch3d}. 

SMPL poses (i.e. 3D joint rotations) and global body rotations are randomly selected from the training splits of UP-3D \cite{Lassner:UP:2017}, 3DPW \cite{vonMarcard2018} and Human3.6M \cite{h36m_pami}. SMPL shapes are obtained in two stages: (i) base body shape coefficients are randomly sampled from $\mathcal{N}(\beta_i; 0, 1.25^2)$ for $i \in \{1, 2, ..., |\boldsymbol{\beta}|\}$, and (ii) measurement offsets from the base body (for each of the 23 measurements listed in Figure \ref{fig:meas_def_table}) are randomly sampled from  $\mathcal{N}(m_j; 0, 0.02^2)$ (units of metres), converted into shape coefficient offsets using the measurements-to-$\beta$s regressor and added to the random base body shape. Step (ii) acts as random body \textit{measurement} augmentation, and is crucial when learning to estimate measurement uncertainties. 

Clothing textures for the SMPL body are randomly selected from SURREAL \cite{varol17_surreal} and MultiGarmentNet \cite{bhatnagar2019mgn}. Background images are obtained from LSUN \cite{yu15lsun}, which contains both indoor and outdoor scenes. Note that background images, intentionally, may contain other humans - this is important for the network to be robust against background humans in real in-the-wild test images.

The sampled SMPL shape, SMPL pose, clothing texture and background image are rendered into a synthetic RGB image using Pytorch3D \cite{ravi2020pytorch3d}. Perspective camera translation is randomly sampled, along with Phong lighting parameters. 2D joint locations (and Gaussian heatmaps) are generated by projecting 3D SMPL joint locations onto the 2D image plane. Synthetic RGB images are converted to edge-images using Canny edge detection \cite{canny1986edge}.

Finally, following \cite{STRAPS2020BMVC, sengupta2021probabilisticposeshape, smith20193dfromsilhouettes}, several data augmentation and corruption methods are applied to the synthetic edge-images and joint heatmaps, to further close the domain gap between training data and noisy test data. Hyperparameters associated with random data generation and augmentation are listed in Table \ref{table:sup_mat_hypparams}.

The synthetic training inputs are paired with ground-truth SMPL pose parameters, body measurements, global body rotations and 2D joint locations, which are each obtained at some point in the synthetic input generation process, as detailed above. Body measurements are computed from sampled SMPL shape coefficients using the $\text{measure}(.)$ operation defined in Section \ref{supmat_sec:body_meas_def}.

\begin{figure}[h!]
    \centering
    \includegraphics[width=\textwidth]{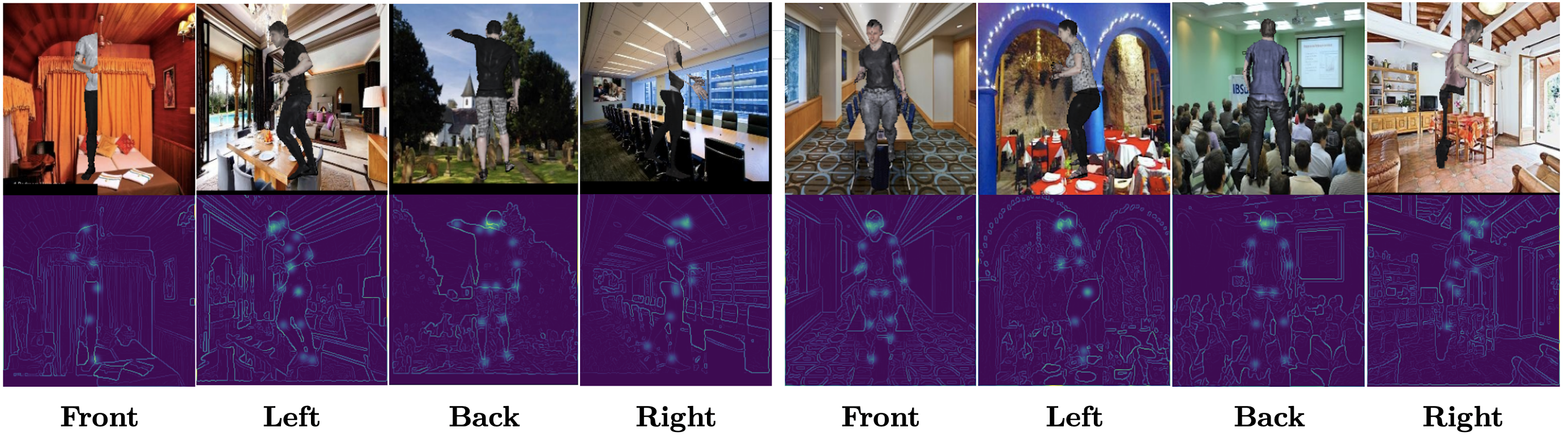}
    \caption{Examples of synthetic RGB evaluation images and corresponding edge-image + 2D joint heatmap proxy representations used for the ablation studies presented in the main manuscript. Each of the 1000 random body shapes in the synthetic evaluation dataset are posed in 4 different configurations, facing forwards, left, backwards and right.}
    \label{fig:synthetic_eval_examples}
\end{figure}

The ablation studies presented in the main manuscript use a synthetic evaluation dataset, which is rendered very similarly to synthetic training data. Examples are given in Figure \ref{fig:synthetic_eval_examples}.

\clearpage
\begin{table}[t!]
\centering
\begin{tabular}{l|c}
    \hline
    \noalign{\smallskip} 
    \textbf{Hyperparameter} & \textbf{Value}\\
    \noalign{\smallskip}
    \hline
    \noalign{\smallskip}
    Camera translation sampling mean & (0, -0.2, 2.5) metres\\
    Camera translation sampling variance & (0.05, 0.05, 0.25) metres\\
    Camera focal length & 300.0\\
    Lighting ambient intensity range & (0.4, 0.8)\\
    Lighting diffuse intensity range & (0.4, 0.8)\\
    Lighting specular intensity range & (0.0, 0.5)\\
    Proxy representation dimensions & $256 \times 256$ pixels\\
    Bounding box scale factor range & (0.8, 1.2)\\
    \multirow{2}{0.6\linewidth}{Body part occlusion probability (divided into 24 DensePose \cite{Guler2018DensePose} parts)} & 0.1 \\
    \\
    \multirow{2}{0.6\linewidth}{2D joints L/R swap probability (for shoulders, elbows, wrists, hips, knees, ankles)} & 0.1\\
    \\
    Vertical/horizontal half occlusion probability & 0.05/0.05\\
    2D joint heatmap removal probability & 0.1\\
    2D joint heatmap location noise range & [-8, 8] pixels\\
    \noalign{\smallskip}
    \hline
    \noalign{\smallskip}
    \end{tabular}
\caption{Hyperparameter values associated with random synthetic training data generation and augmentation.}
\label{table:sup_mat_hypparams}
\end{table}

\begin{table}[b!]
\centering
\begin{tabular}{l|c}
    \hline
    \noalign{\smallskip} 
    \textbf{Method} & \textbf{Single-Image Inference Time (ms)}\\
    \noalign{\smallskip}
    \hline
    \noalign{\smallskip}
    GraphCMR \cite{kolotouros2019cmr} & 33\\
    HMR \cite{hmrKanazawa17} & \textbf{30}\\
    SPIN \cite{kolotouros2019spin} & \textbf{30}\\
    \noalign{\smallskip}
    \hline
    \noalign{\smallskip}
    DaNet \cite{zhang2019danet} & 160\\
    STRAPS \cite{STRAPS2020BMVC} & 250\\
    Sengupta \etal \cite{sengupta2021probabilisticposeshape} & 250\\
    Ours & 140\\
    \noalign{\smallskip}
    \hline
    \noalign{\smallskip}
    \end{tabular}
\caption{Comparison of single-image inference run-times (in milliseconds) for different 3D shape and pose estimation approaches. Methods in the top half do not use proxy representation inputs, while methods in the bottom half do. Proxy representations enable the use of synthetic training data (by closing the synthetic-to-real domain gap), thus overcoming the lack of real training data with accurate and diverse body shape labels. However, proxy representation computation during inference significantly increases run-time. Our approach is faster than recent approaches that use silhouette-based proxy representations \cite{STRAPS2020BMVC, sengupta2021probabilisticposeshape}, since edge-detection is a less-intensive operation than deep-learning-based segmentation. Most (~90\%) of our inference time is due to 2D joint detection \cite{wu2019detectron2}.}
\label{table:sup_mat_runtimes}
\end{table}

\clearpage
\bibliography{egbib}
\end{document}